\documentclass[twoside,leqno,twocolumn]{article}

\usepackage[letterpaper]{geometry}
\usepackage{siamproceedings}

\usepackage[T1]{fontenc}
\usepackage[utf8]{inputenc}
\usepackage{amsfonts}
\usepackage{dsfont}
\usepackage{amsmath}
\usepackage{amsopn}
\usepackage{graphicx}
\usepackage{epstopdf}
\usepackage{enumitem}
\usepackage{algorithm}
\usepackage{algorithmic}
\usepackage{url}
\usepackage{nicefrac}
\usepackage{microtype}
\usepackage{lscape}
\usepackage{fontawesome5}
\usepackage{subcaption}
\usepackage{booktabs}
\usepackage{multirow}
\usepackage{colortbl}
\usepackage[dvipsnames]{xcolor}
\usepackage{pifont}
\usepackage{tikz}
\usepackage[most]{tcolorbox}
\tcbuselibrary{breakable}
\usepackage{listings}
\usepackage{xspace}

\ifpdf
  \DeclareGraphicsExtensions{.eps,.pdf,.png,.jpg}
\else
  \DeclareGraphicsExtensions{.eps}
\fi

\newsiamremark{remark}{Remark}
\newsiamremark{hypothesis}{Hypothesis}
\crefname{hypothesis}{Hypothesis}{Hypotheses}
\newsiamthm{claim}{Claim}

\definecolor{mygray}{gray}{.88}
\definecolor{myblue}{HTML}{5f8fb7}
\definecolor{myred}{HTML}{cc7c7c}
\definecolor{mybrown}{HTML}{D8C6A6}
\definecolor{mossgreen}{HTML}{8B9A7B}
\definecolor{pyblue}{rgb}{0.0, 0.0, 0.5}
\definecolor{pygreen}{rgb}{0.0, 0.5, 0.0}
\definecolor{pyorange}{rgb}{1.0, 0.4, 0.0}

\newcommand{\cmark}{\color{ForestGreen}\ding{51}}%
\newcommand{\xmark}{\color{Red}\ding{55}}%

\newlength\savewidth

\lstdefinestyle{pythonstyle}{
    language=Python,
    basicstyle=\footnotesize\ttfamily,
    breaklines=true,
    morekeywords={self},
    keywordstyle=\color{pyblue},
    commentstyle=\color{pygreen},
    stringstyle=\color{pyorange},
    numberstyle=\tiny\color{gray},
    numbers=left,
    numbersep=10pt,
    tabsize=4,
    showspaces=false,
    showstringspaces=false
}

\newcommand{\method}{{\fontfamily{lmtt}\selectfont \textbf{HopRank}}\xspace}
\newcommand{\llmname}[1]{{\fontfamily{pcr}\selectfont {#1}}\xspace}

\usepackage{amsmath,amsfonts,bm}

\def\eqref#1{equation~\ref{#1}}

\def\1{\bm{1}}

\DeclareMathAlphabet{\mathsfit}{\encodingdefault}{\sfdefault}{m}{sl}
\SetMathAlphabet{\mathsfit}{bold}{\encodingdefault}{\sfdefault}{bx}{n}

\begin{document}

\newcommand\relatedversion{}

\title{\Large \method: Self-Supervised LLM Preference-Tuning on Graphs for Few-Shot Node Classification\relatedversion}
\author{
  \textbf{Ziqing Wang} \quad
  \textbf{Kaize Ding} \\
  Northwestern University \\
  \texttt{ziqingwang2029@u.northwestern.edu} \quad
  \texttt{kaize.ding@northwestern.edu}
}

\date{}

\maketitle


\begin{abstract}
Node classification on text-attributed graphs (TAGs) is a fundamental task with broad applications in citation analysis, social networks, and recommendation systems. Current GNN-based approaches suffer from shallow text encoding and heavy dependence on labeled data, limiting their effectiveness in label-scarce settings. While large language models (LLMs) naturally address the text understanding gap with deep semantic reasoning, existing LLM-for-graph methods either still require abundant labels during training or fail to exploit the rich structural signals freely available in graph topology. Our key observation is that, in many real-world TAGs, edges predominantly connect similar nodes under the homophily principle, meaning graph topology inherently encodes class structure without any labels. Building on this insight, we reformulate node classification as a link prediction task and present \method{}, a fully self-supervised LLM-tuning framework for TAGs. \method{} constructs preference data via hierarchical hop-based sampling and employs adaptive preference learning to prioritize informative training signals without any class labels. At inference, nodes are classified by predicting their connection preferences to labeled anchors, with an adaptive early-exit voting scheme to improve efficiency. Experiments on three TAG benchmarks show that \method{} matches fully-supervised GNNs and substantially outperforms prior graph-LLM methods, despite using zero labeled training data. 
\end{abstract}
\section{Introduction.}
\label{sec: intro}

Node classification on text-attributed graphs (TAGs), where each node carries a rich textual description, is a fundamental task with broad applications in citation analysis, social networks, and recommendation systems~\cite{jin2024large,chen2024exploring}. Graph neural networks (GNNs) have been the dominant paradigm, yet they suffer from two key limitations. First, GNNs encode node text through shallow features such as bag-of-words or fixed embeddings, losing the rich semantics carried by the original text~\cite{he2023harnessing,chen2024llaga}. Second, GNNs require abundant labeled nodes for end-to-end training and degrade sharply in few-shot settings~\cite{kipf2016semi, huang2020graph}. Large language models (LLMs), by contrast, offer deep textual understanding and strong generalization from minimal supervision, capable of few-shot or even zero-shot inference via in-context learning~\cite{ye2024language,sun2023think}. These capabilities have motivated a growing line of work that adapts LLMs for graph tasks~\cite{tang2024graphgpt,chen2024llaga,xu2025gnn,ye2024language, gofa}.

\begin{figure}[t!]
\centering
\includegraphics[height=4.3cm]{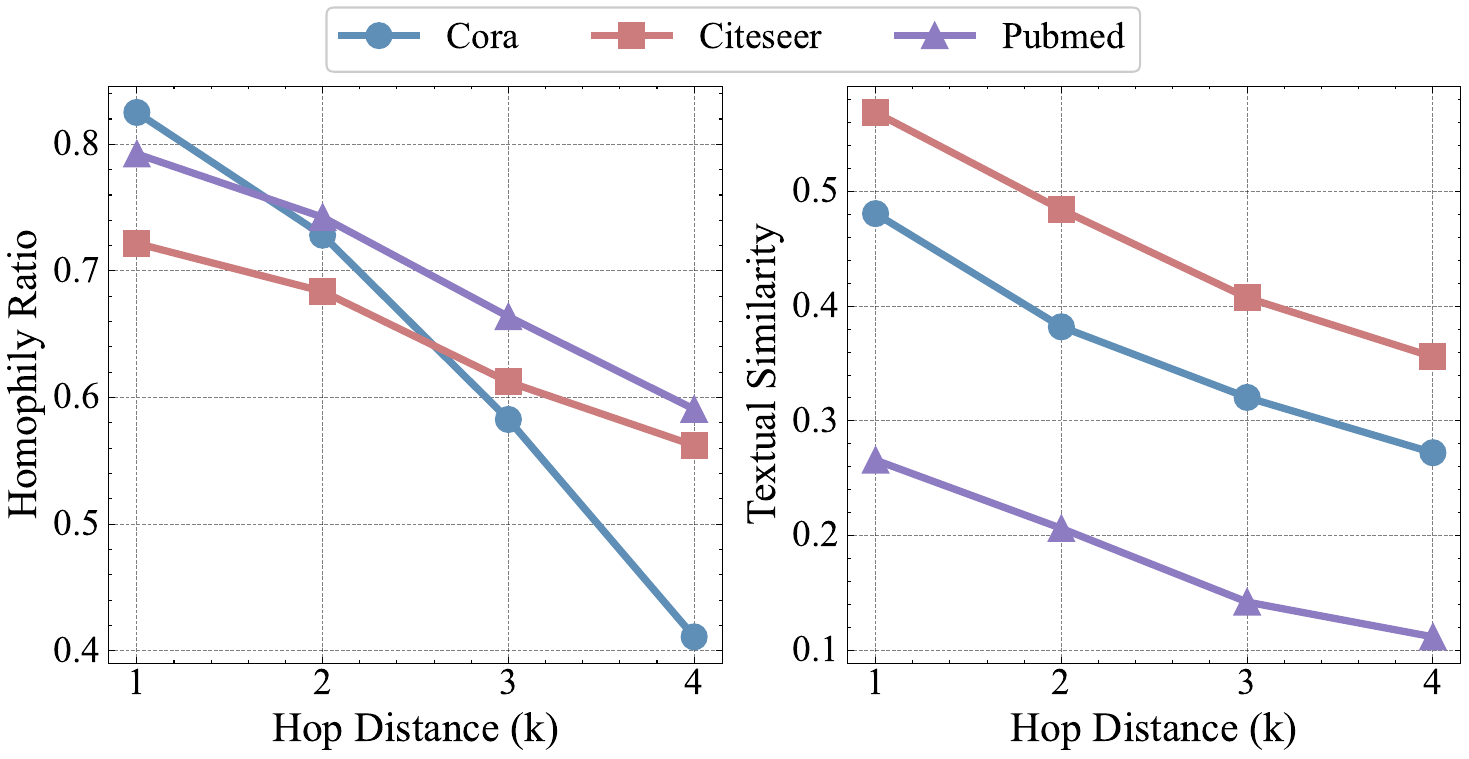}
\caption{\textbf{Graph Topology Encodes Class Structure.} Class homophily and textual similarity are highest for directly connected 
nodes and decay predictably with hop distance.}
\label{fig: homo}
\vspace{-0.3cm}
\end{figure}

However, existing LLM-for-graph methods have yet to fully realize this potential. Prompting-based approaches~\cite{chen2024exploring,wang2024instructgraph} avoid training but achieve limited performance, as they cannot deeply learn the structural patterns encoded in graph topology through in-context learning alone. Instruction-tuning methods~\cite{tang2024graphgpt,ye2024language} and representation-alignment approaches~\cite{chen2024llaga,he2023harnessing} achieve stronger results by fine-tuning on (text, label) pairs or training projection layers, yet they require abundant labeled nodes and degrade in label-scarce settings. Hybrid LLM-GNN frameworks~\cite{glem,gofa} similarly depend on node labels to bridge the language and graph views. Across these paradigms, \emph{the rich structural signals freely encoded in graph topology remain largely unexploited as a source of self-supervision}, leaving LLMs unable to learn class structure without labeled data, even though the graph topology itself may already encode such structure.

To address this gap, we draw on homophily~\cite{mcpherson2001birds}, a well-established principle in network science: similar nodes tend to form connections. This principle governs many prevalent TAG domains, including citation networks, social graphs, and recommendation systems. As Figure~\ref{fig: homo} illustrates, both class homophily and textual similarity are highest for directly connected nodes and decay predictably with hop distance, creating a natural hierarchy: 1-hop neighbors are highly similar, while 3-hop and 4-hop nodes are increasingly dissimilar. This suggests that graph topology inherently encodes a preference ordering over node pairs that reflects the underlying class structure, all without requiring any labels. This raises a key question:

\begin{tcolorbox}[
    enhanced,
    sidebyside, 
    colframe=black!70,
    colback=yellow!5,
    boxrule=1pt, 
    arc=3mm,
    lefthand width=0.08\linewidth,
    sidebyside gap=3mm,
    left=0mm,          
    right=1mm,          
    top=1mm,          
    bottom=1mm,       
    before skip=5pt,    
    after skip=15pt,    
    sidebyside align=center,  
]
\includegraphics[width=1.1\linewidth]{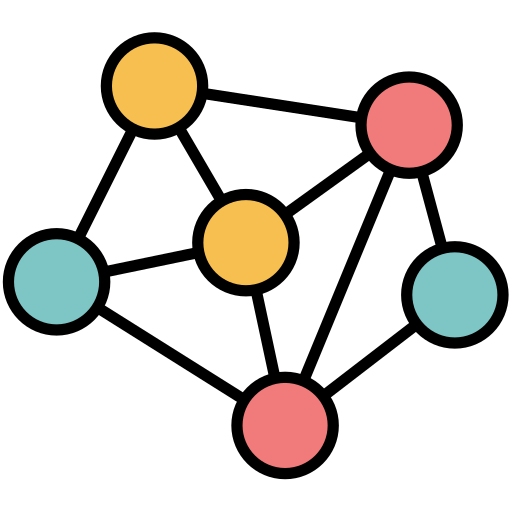}
\tcblower
\small\textit{Can we train LLMs to capture structural preferences from graph topology alone, without any labeled data?}
\end{tcolorbox}

\begin{figure*}[t!]
\centering
\includegraphics[height=5cm]{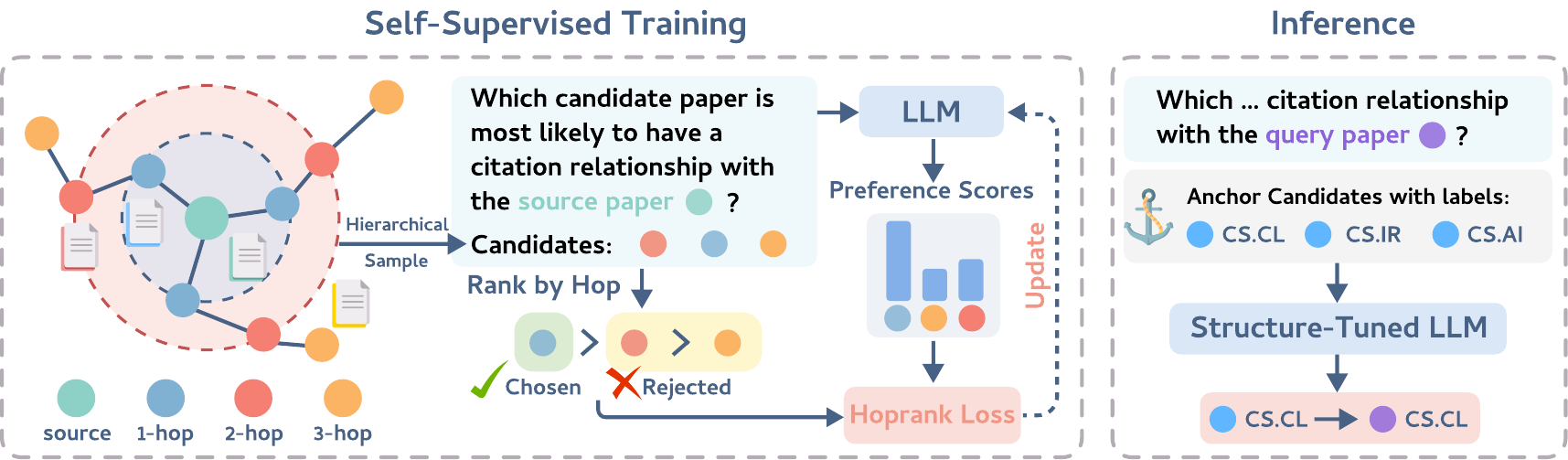}
\caption{\textbf{Overview of the \method{} framework.} \textbf{Left (Self-Supervised Training):} For each edge in the graph, we perform hierarchical hop-based sampling to construct a preference learning instance: the 1-hop neighbor is the \emph{chosen} candidate, while 2-hop and 3-hop nodes serve as \emph{rejected} candidates of decreasing difficulty. The LLM is prompted to identify the most likely connection and trained via the \method{} loss to rank candidates by structural proximity. \textbf{Right (Few-Shot Inference):} Given a query node and one labeled anchor per class, the structure-tuned LLM predicts which anchor the query is most likely connected to, casting classification as a connection preference prediction.}
\label{fig: pipeline}
\end{figure*}

Our key insight is to reformulate node classification as a link prediction task: rather than directly mapping text to class labels, we train the LLM to predict which nodes are most likely to be connected. This perspective shift unlocks a fully self-supervised training paradigm, since the graph topology itself provides all the supervision needed. Because edges in homophilous graphs predominantly connect nodes of the same class, learning to predict connections implicitly teaches the model the similarity patterns that define class membership, without requiring any class labels. Once the model has internalized these connection preferences, few-shot classification follows naturally: given a query node and a small set of labeled anchors per class, the model simply predicts which class's anchors the query node would most likely connect to.

Building on this idea, we introduce \method{}, a framework that trains LLMs through self-supervised learning on graph structure (Figure~\ref{fig: pipeline}). We construct preference data via hierarchical hop-based sampling, where each pair compares an actual 1-hop neighbor (chosen) against nodes sampled from increasing hop distances (rejected). We then employ adaptive preference learning that weights these pairs by (1) hop distance, emphasizing challenging nearby distinctions, and (2) ranking difficulty, focusing on pairs where the model currently makes errors. This creates an automatic curriculum that prioritizes informative training signals without requiring any class labels. At inference, \method{} classifies nodes by predicting their connection preferences to a small set of labeled anchors per class. To further improve efficiency, an adaptive early-exit voting scheme reduces inference cost by up to $\sim\!3.6\times$ with negligible accuracy loss. Extensive experiments show that \method{} matches fully-supervised GNNs and outperforms prior graph-LLM baselines while using zero labeled nodes during training. To summarize, our main contributions are as follows:
\begin{itemize}[leftmargin=*,itemsep=2pt]
    \item \textbf{Framework.} We introduce \method{}, the first self-supervised LLM-tuning framework for TAGs that requires \emph{zero training labels}, reformulating node classification as a link prediction task where graph topology alone supervises the model.

    \item \textbf{Training.} We propose hierarchical hop-based preference sampling coupled with an adaptive DPO objective that weights preference pairs by hop-distance difficulty and current-model ranking errors, turning graph topology into an automatic training curriculum.

    \item \textbf{Efficiency.} We develop an adaptive early-exit voting scheme that reduces inference cost by up to $\sim\!3.6\times$ versus fixed ensemble sampling, with negligible accuracy loss.

    \item \textbf{Effectiveness.} Across three standard TAG benchmarks, \method{} matches fully-supervised GNNs and outperforms prior graph-LLM baselines, while using zero labeled nodes during training.
\end{itemize}
\section{Related Work.}
\label{sec:related}

\paragraph{\textbf{GNN Methods for Node Classification.}}
Graph Neural Networks (GNNs)~\cite{kipf2016semi, hamilton2017inductive, velickovic2017graph, wu2019simplifying} have been the dominant paradigm for node classification, aggregating neighborhood information through message passing with node-level supervised training. While effective under abundant labels, these models treat text attributes as fixed bag-of-words or pre-computed embeddings, losing semantic richness. Several works address few-shot node classification through meta-learning~\cite{zhou2019meta, ding2020more} or prototype-based strategies, yet they still require labeled support sets during both training and inference. \method{} instead harnesses the full textual content through an LLM backbone and replaces label supervision with graph topology, requiring no labeled nodes during training.

\vspace{0.2cm}
\paragraph{\textbf{Self-Supervised Learning on Graphs.}}
Self-supervised graph learning produces node representations without labels, commonly via contrastive objectives~\cite{dgi, grace, bgrl} or masked reconstruction~\cite{graphmae}. These approaches demonstrate that graph structure alone can drive useful representation learning, but they operate on fixed low-dimensional embedding spaces and still require a labeled classifier head for downstream tasks. In contrast, \method{} performs self-supervised fine-tuning of an LLM, turning link prediction into a preference objective that directly prepares the model for few-shot classification without an intermediate feature extraction step.

\vspace{0.2cm}
\paragraph{\textbf{LLMs for Graph Learning.}}
LLMs have demonstrated strong generalization across diverse domains and tasks~\cite{wang2025survey, wang2018remol, wang2025amanda}. For graph learning, recent work integrates LLMs through three main families: (i) prompting-based approaches~\cite{chen2024exploring, wang2024instructgraph} that inject neighbor information into the context, achieving label-free inference but limited performance; (ii) instruction-tuning and representation-alignment methods~\cite{tang2024graphgpt, chen2024llaga, ye2024language, he2023harnessing} that fine-tune on (text, label) pairs for stronger results at the cost of label dependence; and (iii) hybrid LLM-GNN frameworks~\cite{glem, gofa, xu2025gnn} that couple both modalities but still require labeled nodes. \method{} departs from all three families by removing labeled supervision entirely during training: the LLM is tuned only on graph topology, and labels enter the system only at inference as anchors for preference comparison.

\vspace{0.2cm}
\paragraph{\textbf{Preference Learning.}}
Direct Preference Optimization (DPO)~\cite{rafailov2023direct} has emerged as a powerful alternative to reinforcement learning for aligning model outputs with desired preferences, with growing adoption across a wide range of tasks~\cite{wang2025polo, wang2026molmem}. Recent variants such as ORPO~\cite{orpo} and SimPO~\cite{simpo} explore reference-free or SFT-balanced formulations, finding that SFT regularization is critical for stable preference learning. \method{} adapts this paradigm to graph-structured data, introducing hop-distance and ranking-difficulty weights that turn graph topology into an adaptive training curriculum for LLMs.
\section{Method.}
\label{sec: method}

Figure~\ref{fig: pipeline} illustrates the overall \method{} framework, which consists of two stages. In the \emph{self-supervised training} stage (\S\ref{sec:sampling}--\S\ref{sec:preference}), we construct preference data from graph topology via hierarchical hop-based sampling and train the LLM with an adaptive preference objective, requiring no class labels. In the \emph{few-shot inference} stage (\S\ref{sec:inference}), the trained model classifies test nodes by predicting their connection preferences to a small set of labeled anchors, with an adaptive early-exit scheme for efficiency. We begin by formalizing the problem setup.

\subsection{Problem Formulation.}

Let $\mathcal{G} = (\mathcal{V}, \mathcal{E}, \mathbf{X})$ denote a text-attributed graph, where $\mathcal{V} = \{v_1, \ldots, v_N\}$ is the node set, $\mathcal{E} \subseteq \mathcal{V} \times \mathcal{V}$ is the edge set, and $\mathbf{X} = \{x_1, \ldots, x_N\}$ assigns a textual description $x_i$ to each node $v_i$. Each node belongs to one of $C$ classes, $y_i \in \mathcal{C} = \{c_1, \ldots, c_C\}$. In the standard node classification setting, the goal is to learn a mapping $g: x_i \rightarrow y_i$ using a set of labeled nodes $\mathcal{V}_L \subset \mathcal{V}$.

We reformulate this task as a link prediction problem. Instead of learning $g$ directly, we train the LLM to learn a connection preference function:
\begin{align}
f_\theta: (x_u, x_v) \rightarrow [0, 1]
\end{align}
which estimates the likelihood that two nodes $(u, v)$ are connected, given only their textual descriptions and the graph topology $\mathcal{E}$. Crucially, this function is learned in a fully self-supervised manner, requiring no class labels. At inference, classification reduces to predicting which class's labeled anchors a test node would most likely connect to:
\begin{align}
\hat{y}_t = \arg\max_{c \in \mathcal{C}} \; f_\theta(x_{v_t}, x_{a^c})
\end{align}
where $\mathcal{A}_c = \{a_1^c, \ldots, a_K^c\}$ are $K$ labeled anchor nodes per class $c$.

\subsection{Hierarchical Hop-based Sampling.}
\label{sec:sampling}

To leverage the structural preferences encoded in graph topology, we construct training instances through hierarchical hop-based sampling. For a source node $u$, let $\mathcal{N}_h(u)$ denote the set of nodes reachable in exactly $h$ hops but not fewer. For each edge $(u, v) \in \mathcal{E}$, we create a preference learning instance with the true neighbor $v \in \mathcal{N}_1(u)$ as the \emph{chosen} response and negative candidates sampled from increasing hop distances as \emph{rejected} responses:
\begin{align}
n_h \sim \mathcal{N}_h(u), \quad h \in \{2, 3, \ldots, k\}
\end{align}
where $k$ is the maximum hop budget. This creates a natural curriculum: 2-hop negatives are challenging (semantically similar but structurally unconnected), while distant negatives are progressively easier to distinguish, as validated by the decay patterns in Figure~\ref{fig: homo}. In practice, we collect candidates via a bounded local BFS of depth $k$ from each source node, keeping the per-edge cost at $\mathcal{O}(\bar{d}^{\,k})$ where $\bar{d}$ is the average degree.

To avoid positional shortcuts, the chosen and rejected candidates are \emph{randomly shuffled} across the candidate slots (A/B/C/\ldots) for every training instance, so the model cannot exploit the order of presentation as a cue. We format each instance as a natural language prompt. For example, with two negative samples:

\begin{tcolorbox}[colback=myblue!5!white, colframe=myblue!85!black, title=Link Prediction Training Prompt]
\small
\textbf{\texttt{Given a source paper and multiple candidate papers, identify which candidate paper is most likely to have a citation relationship with the source paper.}}

\texttt{Source Paper: '\{title\_u\}'}\\
\texttt{Abstract: \{abstract\_u\}}

\texttt{Candidate Papers:}\\
\texttt{A. '\{title\_v\}'}\\
\texttt{Abstract: \{abstract\_v\}} \textcolor{myred}{[1-hop: actual neighbor]}

\texttt{B. '\{title\_n2\}'}\\
\texttt{Abstract: \{abstract\_n2\}} \textcolor{mybrown}{[2-hop: hard negative]}

\texttt{C. '\{title\_n3\}'}\\
\texttt{Abstract: \{abstract\_n3\}} \textcolor{mossgreen}{[3-hop: easy negative]}

\textbf{\texttt{Which candidate paper is most likely to be cited by or cite the source paper?}}
\end{tcolorbox}
\vspace{0.3cm}

\subsection{Adaptive Preference Learning.}
\label{sec:preference}

After hierarchical sampling, we obtain a listwise preference dataset where each instance contains one chosen response (1-hop neighbor) and multiple rejected responses at varying distances. We train the LLM to predict these connections by learning which pairs are most informative.

Not all preference pairs are equally useful. Near-hop negatives (e.g., 2-hop) are semantically similar to the chosen neighbor yet structurally unconnected, making them challenging and informative, while distant negatives (e.g., 5-hop) are trivially distinguishable. Moreover, the model's ranking quality evolves during training, so different pairs deserve different emphasis at different stages. To address this, we introduce adaptive preference learning through two complementary weighting mechanisms.

We build upon Direct Preference Optimization (DPO)~\cite{rafailov2023direct}, which parameterizes preferences through implicit rewards:
\begin{align}
\psi(y|x) = \beta \log\frac{\pi_\theta(y|x)}{\pi_{\text{ref}}(y|x)}
\end{align}
where $\psi(y|x)$ measures the log-likelihood ratio of response $y$ under the current policy $\pi_\theta$ versus the reference policy $\pi_{\text{ref}}$, scaled by $\beta$. The base DPO loss for a chosen-rejected pair $(y_c, y_r)$ is:
\begin{align}
\mathcal{L}_{\text{DPO}}(y_c, y_r) = -\log \sigma(\psi(y_c|x) - \psi(y_r|x))
\end{align} 

We extend DPO with two complementary weights that implement adaptive preference learning:

\textbf{Distance weight} $w_{\text{dist}}$ emphasizes challenging nearby distinctions:
\begin{align}
w_{\text{dist}}(y_c, y_r) = \frac{1}{|d(y_c) - d(y_r)|}
\end{align}
where $d(\cdot)$ denotes hop distance from the source node (with $d(y_c)=1$ for the chosen neighbor). The weight is maximized for 2-hop negatives ($w=1$) and decreases for distant negatives (e.g., $w=0.25$ for 5-hop).

\textbf{Ranking weight} $w_{\text{rank}}$ focuses on the model's current errors:
\begin{align}
w_{\text{rank}}(y_c, y_r) = \left|\frac{1}{\text{rank}(y_c)} - \frac{1}{\text{rank}(y_r)}\right|
\end{align}
where $\text{rank}(y)$ is the position when all candidates are sorted by $\psi(y|x)$ in descending order. The reciprocal weighting emphasizes top-position errors: if the model incorrectly ranks a negative above the true neighbor, this pair receives higher weight, focusing learning on errors most critical for downstream classification.

\textbf{Final objective} combines weighted preference learning with SFT regularization:
\begin{align}
\label{eq:hoprank}
\mathcal{L}_{\text{HopRank}} = &\sum_{(y_c, y_r) \in \mathcal{P}} w_{\text{dist}}(y_c, y_r) \cdot w_{\text{rank}}(y_c, y_r) \nonumber\\
&\quad \cdot \mathcal{L}_{\text{DPO}}(y_c, y_r) + \gamma \cdot \mathcal{L}_{\text{SFT}}(y_c)
\end{align}
where $\mathcal{P}$ contains all chosen-rejected pairs from our listwise preferences and $\mathcal{L}_{\text{SFT}}(y_c) = -\log \pi_\theta(y_c|x)$. The two components play complementary roles: the SFT term maintains domain grounding and correct-target supervision, while the adaptive DPO terms provide the discriminative signal that separates true neighbors from structurally close negatives. In practice, a large $\gamma$ (we use $\gamma{=}5.0$) is necessary because the SFT gradient decays rapidly once the model learns to generate neighbor titles, a phenomenon consistent with findings in ORPO~\cite{orpo} and SimPO~\cite{simpo}. Our ablation (\S\ref{sec: exp}) confirms that removing any of $w_{\text{dist}}$, $w_{\text{rank}}$, or $\mathcal{L}_{\text{SFT}}$ individually degrades accuracy.

\subsection{Few-Shot Inference via Anchor Sampling.}
\label{sec:inference}

During inference, we leverage the learned preference model for few-shot classification. Given $K$ labeled anchor nodes per class $\mathcal{A}_c = \{a_1^c, \ldots, a_K^c\}$, directly comparing a test node against all anchors in one step is impractical due to context length constraints when $K$ is large. Instead, we employ a sampling-based voting strategy.

In each round $t$, we randomly sample one anchor $a^c$ from each class and classify the test node $v_t$ by predicting which sampled anchor it would most likely connect to via $f_\theta$. After repeating this $R$ times with different anchor combinations, the final class is determined via majority voting:
\begin{align}
\hat{y}_t = \arg\max_{c \in \mathcal{C}} \sum_{r=1}^{R} \mathbf{1}[\hat{c}_r = c]
\end{align}
where $\hat{c}_r$ is the predicted class in round $r$. The number of rounds $R$ can be flexibly set: when $R = K$, each anchor is used once (standard few-shot setting); when $R > K$, \textbf{ensemble sampling} with replacement allows each anchor to be evaluated multiple times in different contexts. This ensemble approach reduces variance and improves classification reliability.

\paragraph{\textbf{Adaptive Early Exit.}}
Under a fixed budget, the model spends equal compute on confident and uncertain nodes alike. We introduce an \emph{adaptive early-exit} strategy that checkpoints the vote tally every $\Delta$ rounds and terminates as soon as any class accumulates more than a majority-defining fraction $\tau$ of the votes cast so far. Formally, after $t$ rounds the procedure exits if
\begin{align}
\max_{c \in \mathcal{C}} \; \frac{\text{votes}_c(t)}{t} \; > \; \tau,
\end{align}
otherwise voting continues until $t = R_{\max}$. Because structurally consistent anchors tend to produce consistent predictions, many test nodes pass the threshold well before the full budget is exhausted, yielding significant speedups with negligible accuracy loss (\S\ref{sec: exp}). The scheme degrades gracefully: on hard-to-classify nodes it naturally reverts to the full fixed-budget ensemble.

\subsection{Algorithm Summary and Complexity.}
\label{sec:algorithm}

Algorithm~\ref{alg:hoprank} summarizes the full \method{} pipeline. The training stage is fully self-supervised; labels enter the system only as inference-time anchors.

\begin{algorithm}[t!]
\caption{\method{} Pipeline}
\label{alg:hoprank}
\small
\begin{algorithmic}
\STATE \textbf{Input:} TAG $\mathcal{G}{=}(\mathcal{V},\mathcal{E},\mathbf{X})$, LLM $\pi_\theta$, anchors $\{\mathcal{A}_c\}_{c\in\mathcal{C}}$
 
\STATE
\STATE {\color{Purple}\textsc{Stage 1: Data Construction}} {\color{ForestGreen}\textit{(offline, no labels)}}
\FOR {each $(u,v)\!\in\!\mathcal{E}$}
  \STATE Sample negatives $\{n_h\}_{h=2}^{k}$ via BFS from $u$
  \STATE Shuffle slots; emit prompt with $v$ and $\{n_h\}$
\ENDFOR
 
\STATE
\STATE {\color{Purple}\textsc{Stage 2: Preference Tuning}} {\color{ForestGreen}\textit{(self-supervised)}}
\FOR {each instance $(x, y_c, \{y_r\})$}
  \STATE Compute $\psi(y|x)$, weights $w_{\text{dist}}$, $w_{\text{rank}}$
  \STATE Update $\pi_\theta$: minimize $\mathcal{L}_{\text{HopRank}}$ (Eq.\,\ref{eq:hoprank})
\ENDFOR
 
\STATE
\STATE {\color{Purple}\textsc{Stage 3: Inference}} {\color{ForestGreen}\textit{(few-shot, early exit)}}
\STATE $\text{votes}_c \gets 0\;\;\forall c$; $\;t\gets 0$
\WHILE{$t < R_{\max}$}
  \STATE Sample $a^c\!\sim\!\mathcal{A}_c$ per class; $\;\hat{c} \gets f_\theta(v_t, \{a^c\})$
  \STATE $\text{votes}_{\hat{c}} \mathrel{+}= 1$; $\;t\gets t+1$
  \IF{$t \bmod \Delta = 0$ \AND $\max_c \frac{\text{votes}_c}{t} > \tau$}
    \STATE \textbf{break}
  \ENDIF
\ENDWHILE
\RETURN $\arg\max_{c}\;\text{votes}_c$
\end{algorithmic}
\end{algorithm}

\paragraph{\textbf{Complexity Analysis.}}
\label{sec: complexity}
The preference-data construction is a \emph{one-time offline} step. For each edge $(u,v) \in \mathcal{E}$, a bounded local BFS of depth $k$ collects hop-indexed negatives at cost $\mathcal{O}(\bar{d}^{\,k})$ per edge, where $\bar{d}$ is the average degree. The total preprocessing time is $\mathcal{O}(|\mathcal{E}| \cdot \bar{d}^{\,k})$, linear in the number of edges for bounded $\bar{d}$ and $k$, finishing in minutes on all benchmarks. Memory requires only the sparse adjacency and preference lists, yielding $\mathcal{O}(|\mathcal{V}|+|\mathcal{E}|)$. Training inherits the cost of standard LoRA-based LLM fine-tuning, and inference scales linearly in the number of voting rounds $R$, further reduced by adaptive early exit (\S\ref{sec: exp}).
\section{Experiments.}
\label{sec: exp}

\begin{table*}[t!]
\centering
\caption{\textbf{Main Results on Node Classification Benchmarks.} Accuracy (\%) for few-shot learning on three TAGs. $K$-shot denotes the number of labeled anchors per class used at inference. ``\#Lbl?'' indicates whether the method requires labeled nodes during \emph{training}. Best in \textbf{bold}.}
\label{tab:main_results}
\renewcommand{\arraystretch}{1.2}  
\resizebox{\textwidth}{!}{
\begin{tabular}{l c c c c c c c c c c c c}
\toprule
\multirow{2}{*}{\textbf{Method}} & \multirow{2}{*}{\textbf{\#Lbl?}} & \multicolumn{3}{c}{\textbf{Citeseer}} && \multicolumn{3}{c}{\textbf{Cora}} && \multicolumn{3}{c}{\textbf{Pubmed}} \\
\cmidrule{3-5} \cmidrule{7-9} \cmidrule{11-13}
& & 5-shot & 10-shot & 20-shot && 5-shot & 10-shot & 20-shot && 5-shot & 10-shot & 20-shot \\
\midrule
\rowcolor{gray!15}
\multicolumn{13}{c}{\textit{\textbf{Non-LLM Baselines}}} \\
\llmname{GCN}       & \cmark & 63.21$_{\pm 0.87}$ & 68.38$_{\pm 1.49}$ & 69.75$_{\pm 0.42}$ && \textbf{71.72}$_{\pm 0.22}$ & \underline{78.22}$_{\pm 0.89}$ & \underline{81.30}$_{\pm 0.59}$ && 70.58$_{\pm 0.49}$ & 75.33$_{\pm 0.94}$ & 79.64$_{\pm 0.98}$ \\
\llmname{GraphSAGE} & \cmark & 62.35$_{\pm 0.83}$ & 67.02$_{\pm 0.56}$ & 69.95$_{\pm 0.66}$ && 70.66$_{\pm 1.09}$ & 77.98$_{\pm 0.32}$ & \textbf{81.37}$_{\pm 0.43}$ && 68.59$_{\pm 1.01}$ & 75.12$_{\pm 0.55}$ & 77.64$_{\pm 0.92}$ \\
\llmname{SGC}       & \cmark & 62.08$_{\pm 0.77}$ & 67.44$_{\pm 0.60}$ & \underline{70.92}$_{\pm 0.77}$ && 71.48$_{\pm 0.35}$ & \textbf{78.44}$_{\pm 0.37}$ & 79.84$_{\pm 0.68}$ && 70.74$_{\pm 0.94}$ & 74.98$_{\pm 1.91}$ & 78.45$_{\pm 0.56}$ \\
\llmname{GAT}       & \cmark & 63.48$_{\pm 1.54}$ & 68.10$_{\pm 1.77}$ & 70.02$_{\pm 0.44}$ && \underline{71.70}$_{\pm 0.57}$ & 77.58$_{\pm 0.23}$ & 80.94$_{\pm 0.56}$ && 69.56$_{\pm 1.55}$ & 72.84$_{\pm 1.36}$ & 78.37$_{\pm 0.38}$ \\
\midrule
\rowcolor{gray!15}
\multicolumn{13}{c}{\textit{\textbf{LLM-based Baselines (LLaMA-3-8B)}}} \\
Zero-Shot & \xmark & 58.17$_{\pm 0.64}$ & 58.17$_{\pm 0.64}$ & 58.17$_{\pm 0.64}$ && 65.54$_{\pm 0.26}$ & 65.54$_{\pm 0.26}$ & 65.54$_{\pm 0.26}$ && 74.51$_{\pm 0.39}$ & 74.51$_{\pm 0.39}$ & 74.51$_{\pm 0.39}$ \\
w/\,Neighbors & \xmark & 54.93$_{\pm 1.28}$ & 54.93$_{\pm 1.28}$ & 54.93$_{\pm 1.28}$ && 68.72$_{\pm 1.56}$ & 68.72$_{\pm 1.56}$ & 68.72$_{\pm 1.56}$ && 74.98$_{\pm 3.16}$ & 74.98$_{\pm 3.16}$ & 74.98$_{\pm 3.16}$ \\
w/\,Few-Shot & \cmark & 62.40$_{\pm 1.47}$ & 62.92$_{\pm 1.32}$ & 63.28$_{\pm 1.50}$ && 64.50$_{\pm 2.40}$ & 65.57$_{\pm 2.13}$ & 67.86$_{\pm 2.22}$ && 77.12$_{\pm 1.34}$ & 79.75$_{\pm 1.12}$ & 81.97$_{\pm 1.25}$ \\
\llmname{InstructTuning} & \cmark & \underline{68.77}$_{\pm 2.66}$ & \underline{69.10}$_{\pm 1.63}$ & 68.44$_{\pm 0.93}$ && 69.33$_{\pm 1.50}$ & 69.53$_{\pm 1.21}$ & 71.89$_{\pm 1.76}$ && \underline{83.71}$_{\pm 0.64}$ & \underline{85.03}$_{\pm 0.16}$ & \underline{88.16}$_{\pm 2.17}$ \\
\llmname{LLaGA}     & \cmark & 43.71$_{\pm 4.36}$ & 51.22$_{\pm 1.43}$ & 66.17$_{\pm 0.43}$ && 62.88$_{\pm 2.19}$ & 69.25$_{\pm 0.97}$ & 76.51$_{\pm 0.22}$ && 58.63$_{\pm 1.05}$ & 67.29$_{\pm 2.26}$ & 68.77$_{\pm 0.35}$ \\
\llmname{GraphGPT}  & \cmark & 51.83$_{\pm 2.24}$ & 55.40$_{\pm 3.16}$ & 62.84$_{\pm 0.49}$ && 60.17$_{\pm 1.44}$ & 61.58$_{\pm 0.77}$ & 63.25$_{\pm 0.34}$ && 57.39$_{\pm 3.67}$ & 71.33$_{\pm 2.81}$ & 78.36$_{\pm 1.24}$ \\
\midrule
\rowcolor{myred!20}
\textbf{\method (Ours)} & \xmark & \textbf{69.03}$_{\pm 1.11}$ & \textbf{71.31}$_{\pm 1.06}$ & \textbf{72.91}$_{\pm 1.02}$ && 70.37$_{\pm 1.42}$ & 73.73$_{\pm 1.33}$ & 77.56$_{\pm 1.08}$ && \textbf{88.00}$_{\pm 1.97}$ & \textbf{88.62}$_{\pm 1.24}$ & \textbf{89.98}$_{\pm 0.60}$ \\
\bottomrule
\end{tabular}
}

\end{table*}

We evaluate \method{} on three standard text-attributed citation benchmarks to assess its effectiveness, efficiency, and robustness. Our experiments are designed to answer four research questions:
 
\begin{itemize}[leftmargin=*, itemsep=1pt, topsep=3pt]
    \item[$\spadesuit$] \textbf{RQ1:} Can self-supervised preference tuning on graph topology match or outperform supervised methods for few-shot node classification? (\S\ref{sec:main})
    \item[$\spadesuit$] \textbf{RQ2:} Can adaptive early exit reduce inference cost without sacrificing accuracy? (\S\ref{sec:efficiency})
    \item[$\spadesuit$] \textbf{RQ3:} How sensitive is \method{} to its key design choices? (\S\ref{sec:analysis})
    \item[$\spadesuit$] \textbf{RQ4:} How does each component of \method{} contribute to overall performance? (\S\ref{sec:ablation})
\end{itemize}

\subsection{Experimental Setup.}
\label{sec:setup}
 
In this section, we describe the experimental setup, including the datasets, evaluation protocol, baselines, and implementation details.

 \vspace{0.3cm}
\paragraph{\textbf{Datasets.}}
We evaluate \method{} on three standard text-attributed citation benchmarks: Cora~\cite{sen2008collective}, Citeseer~\cite{giles1998citeseer}, and Pubmed~\cite{yang2016revisiting}. These datasets span a meaningful range of homophily ratios (72.9\%--82.5\%), allowing us to probe robustness within the homophilous regime targeted by our design. Key statistics are in Table~\ref{tab:dataset_stats}; full descriptions are in Appendix~\ref{app:datasets}.

\begin{table}[h!]
\centering
\caption{\textbf{Summary of evaluation benchmarks.} Homophily ratio is the fraction of edges connecting nodes of the same class.}
\label{tab:dataset_stats}
\renewcommand{\arraystretch}{1.2} 
\resizebox{\linewidth}{!}{
\begin{tabular}{lcccc}
\toprule
\textbf{Dataset} & \textbf{Nodes} & \textbf{Edges} & \textbf{Classes} & \textbf{Homophily (\%)} \\
\midrule
Cora      & 2{,}708 & 5{,}278  & 7 & 82.5 \\
Citeseer  & 3{,}327 & 4{,}676  & 6 & 72.9 \\
Pubmed    & 19{,}717& 44{,}324 & 3 & 79.2 \\
\bottomrule
\end{tabular}}
\end{table}

\vspace{0.3cm}
\paragraph{\textbf{Evaluation protocol.}}
For every dataset we use the standard few-shot evaluation setup with $K \in \{5, 10, 20\}$ labeled nodes per class. For \method{}, these labels are used \emph{only at inference} as anchors; no class labels appear in the training signal. During inference, we use $R{=}100$ voting rounds by default, with the adaptive early-exit strategy (checkpoint interval $\Delta{=}10$, threshold $\tau{=}0.5$) applied in the efficiency experiments (\S\ref{sec:efficiency}). All results are averaged over 5 runs with different random seeds; standard deviations are reported as subscripts.

\vspace{0.3cm}
\paragraph{\textbf{Baselines.}}
\ding{182} \emph{\textbf{Non-LLM baselines.}} We compare against four representative GNN backbones trained end-to-end with node-level supervision: GCN~\cite{kipf2016semi}, GraphSAGE~\cite{hamilton2017inductive}, SGC~\cite{wu2019simplifying}, and GAT~\cite{velickovic2017graph}. All GNN models use the same $K$ labeled nodes per class for supervised training.
\ding{183} \emph{\textbf{LLM-based baselines.}} We evaluate six methods built on LLaMA-3-8B, covering the three paradigms discussed in \S\ref{sec: intro}. \emph{Prompting-based:} Zero-Shot (text only), w/ Neighbors (text + 1-hop neighbor titles), and w/ Few-Shot (in-context learning with $K$ demonstrations per class). \emph{Tuning-based:} InstructTuning~\cite{ouyang2022training} (LoRA fine-tuning on $K$ labeled nodes). \emph{Graph-LLM:} LLaGA~\cite{chen2024llaga} (GNN-to-LLM projection) and GraphGPT~\cite{tang2024graphgpt} (graph instruction tuning).

\vspace{0.3cm}
\paragraph{\textbf{Implementation details.}}
\method{} uses LLaMA-3-8B-Instruct as the backbone with LoRA (rank 8, alpha 16) applied to all linear layers. We set the DPO temperature $\beta{=}0.1$, SFT weight $\gamma{=}5.0$, and hop budget $k{=}3$ (sampling from 2-hop and 3-hop neighborhoods). Training uses AdamW with learning rate $2\!\times\!10^{-4}$ and cosine annealing.

\begin{figure*}[t!]
\centering
\includegraphics[width=\textwidth]{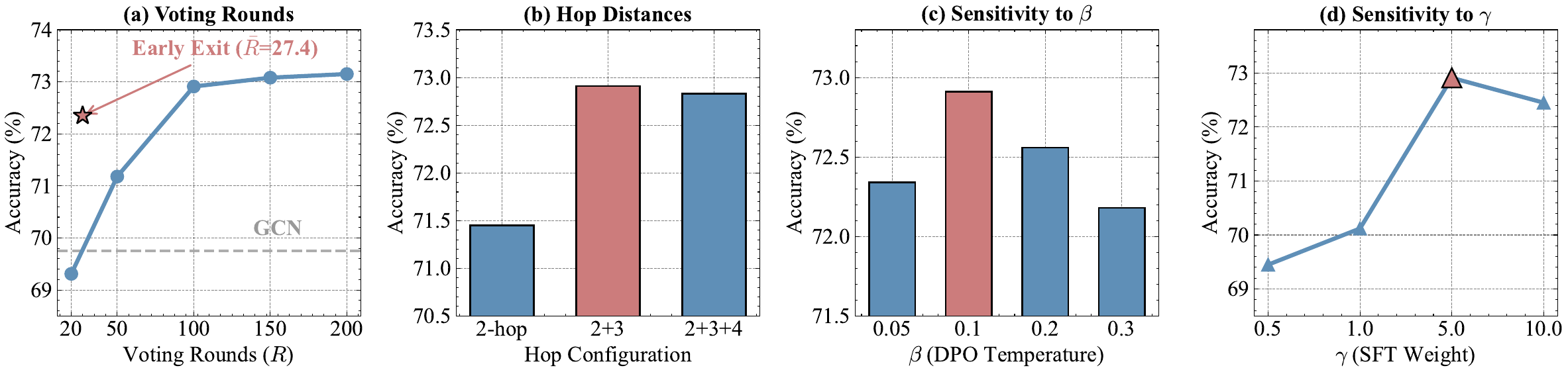}
\caption{\textbf{Analysis of Design Choices on Citeseer-20.} (a) Accuracy improves rapidly with voting rounds $R$ and plateaus beyond $R{=}100$; the adaptive early-exit ($\star$) achieves comparable accuracy at $\bar{R}{=}27.4$ rounds. (b) 2-hop + 3-hop is the optimal hop configuration; adding 4-hop dilutes training with trivially-distinct pairs. (c) \method{} is robust to DPO temperature $\beta$. (d) SFT weight $\gamma$ is more sensitive: small values cause the SFT gradient to vanish early. Red markers indicate default settings.}
\label{fig:analysis}
\end{figure*}

\subsection{Main Results (\texorpdfstring{$\spadesuit$}{♠} RQ1).}
\label{sec:main}

Table~\ref{tab:main_results} summarizes the overall comparison. \method{} achieves the best accuracy across all few-shot settings on Citeseer and Pubmed, and remains competitive on Cora, despite using zero labeled nodes during training. On Pubmed, \method{} attains 88.00\%, 88.62\%, and 89.98\% across the three shot regimes, surpassing the strongest fully-supervised GNN (GCN at 79.64\% on 20-shot) by more than 10 points. On Citeseer, \method{} consistently outperforms all baselines, reaching 72.91\% at 20-shot. On Cora, \method{} is competitive at $K{=}20$ (77.56\%) but trails the best GNNs at $K{=}5$, as the small anchor budget limits voting reliability; this gap narrows as $K$ grows and the ensemble increasingly amortizes anchor noise.

\vspace{0.3cm}
\paragraph{\textbf{Comparison with LLM-based methods.}}
Among LLM-based approaches, InstructTuning is the strongest baseline, achieving 88.16\% on Pubmed-20 with supervised fine-tuning. \method{} surpasses it (89.98\%) without using any training labels, demonstrating that self-supervised structural signals can be more effective than label-dependent tuning. LLaGA and GraphGPT, despite using labeled supervision, underperform \method{} by 8--25 points on Citeseer and Pubmed in the 5-shot setting. This confirms that label-efficient fine-tuning on a small labeled set is insufficient to teach the LLM the connection patterns of the graph, while \method{}'s hop-based preference tuning captures these patterns without any labels.

\begin{table*}[t!]
    \centering
\caption{\textbf{Ablation Study of \method{} Components.} Accuracy (\%) on all three benchmarks at 20-shot. Each row removes one component from the full model.}
\label{tab:ablation_study}
    \renewcommand{\arraystretch}{1.2} 
    \resizebox{0.6\linewidth}{!}{%
        \begin{tabular}{l c c c}
        \toprule
        \textbf{Model Configuration} & \textbf{Citeseer} & \textbf{Cora} & \textbf{Pubmed} \\
        \midrule
        \rowcolor{gray!15}
        \textbf{\method{} (Full Model)} & \textbf{72.91}$_{\pm 1.02}$ & \textbf{77.56}$_{\pm 1.08}$ & \textbf{89.98}$_{\pm 0.60}$ \\
        \midrule
        \multicolumn{4}{c}{\textit{--- Ablation of Core Components ---}} \\
        w/o Hierarchical Sampling        & 70.19$_{\pm 1.54}$ & 74.42$_{\pm 1.50}$ & 87.52$_{\pm 1.41}$ \\
        w/o Self-supervised Training     & 38.71$_{\pm 2.07}$ & 51.51$_{\pm 3.80}$ & 42.15$_{\pm 2.22}$ \\
        w/o Ensemble Sampling ($R{=}K$) & 69.31$_{\pm 1.94}$ & 71.36$_{\pm 2.55}$ & 87.62$_{\pm 1.24}$ \\
        \midrule
        \multicolumn{4}{c}{\textit{--- Ablation of Loss Components ---}} \\
        w/o Distance Weight ($w_{\text{dist}}$) & 72.19$_{\pm 1.34}$ & 74.79$_{\pm 1.49}$ & 88.10$_{\pm 0.87}$ \\
        w/o Rank Weight ($w_{\text{rank}}$)     & 71.63$_{\pm 1.45}$ & 75.16$_{\pm 1.27}$ & 88.95$_{\pm 0.93}$ \\
        w/o SFT ($\mathcal{L}_{\text{SFT}}$)     & 68.87$_{\pm 1.30}$ & 73.25$_{\pm 1.33}$ & 86.33$_{\pm 0.62}$ \\
        \bottomrule
        \end{tabular}
    }
\end{table*}

\subsection{Efficiency Analysis (\texorpdfstring{$\spadesuit$}{} RQ2).}
\label{sec:efficiency}

Our ensemble voting is the dominant inference cost, scaling linearly in the number of rounds $R$. We treat $R$ as a proxy for wall-clock compute. As shown in Figure~\ref{fig:analysis}(a), accuracy rises rapidly from the baseline budget $R{=}K{=}20$ (69.31\%) to the full ensemble $R{=}100$ (72.91\%), then plateaus. To avoid spending the full budget on every node, we activate the adaptive early-exit strategy of \S\ref{sec:inference} (checkpoint $\Delta{=}10$, threshold $\tau{=}0.5$, cap $R_{\max}{=}100$).

Adaptive early exit ($\star$ in Figure~\ref{fig:analysis}(a)) recovers the accuracy of the full $R{=}100$ ensemble within 0.6 accuracy points (72.35\%) while using only 27.4 rounds on average, a $\sim\!3.6\times$ reduction over $R{=}100$ and only a $1.37\times$ overhead over the minimal $R{=}20$ baseline. The effectiveness of early exit stems from a key property of our framework: when the learned preference model is confident about a node's class, different random anchor combinations produce consistent predictions, allowing the vote to converge quickly. Compute is thus automatically reallocated from easy nodes to genuinely ambiguous cases. Combined with the one-time nature of offline preference data construction (\S\ref{sec:algorithm}), the end-to-end compute profile of \method{} is dominated by LLM fine-tuning itself, not by our auxiliary pipeline.

\subsection{Analysis of Design Choices (\texorpdfstring{$\spadesuit$}{} RQ3).}
\label{sec:analysis}

We examine the key design choices that instantiate our framework. Figure~\ref{fig:analysis} summarizes all results.

\vspace{0.3cm}
\paragraph{\textbf{Impact of Voting Rounds ($R$).}}
As shown in Figure~\ref{fig:analysis}(a), accuracy rises rapidly from 69.31\% at $R{=}K{=}20$ to 72.91\% at $R{=}100$, then plateaus with gains under 0.25\% beyond $R{=}100$. This behavior reflects a variance reduction effect analogous to bootstrap aggregation: each round samples a different anchor combination, and averaging over more combinations reduces prediction variance. The clear saturation at $R{=}100$ indicates that the preference model has extracted all available signal from the anchor pool, motivating $R{=}100$ as our default budget.

\vspace{0.3cm}
\paragraph{\textbf{Impact of Hop Distances.}}
Figure~\ref{fig:analysis}(b) shows that combining 2-hop and 3-hop negatives achieves the best accuracy of 72.91\%. Using only 2-hop negatives yields 71.45\%, suggesting that the model benefits from a curriculum of varying difficulty levels rather than exclusively hard negatives. This mirrors findings in contrastive learning, where a mix of hard and moderate negatives produces better representations than hard negatives alone. Adding 4-hop negatives slightly reduces accuracy to 72.83\%, as they dilute the training budget with trivially-distinct pairs that $w_{\text{dist}} = 1/|d(y_c) - d(y_r)|$ already downweights.

\vspace{0.3cm}
\paragraph{\textbf{Hyperparameter Sensitivity.}}
Figures~\ref{fig:analysis}(c) and (d) vary $\beta$ and $\gamma$ independently. \method{} is robust to the DPO temperature $\beta$: accuracy stays within a 0.73\% range across $\beta \in [0.05, 0.3]$, indicating that the preference ordering induced by hop distance is sufficiently clear that the model learns it across a wide range of temperatures. The SFT weight $\gamma$ is more sensitive: at $\gamma{=}0.5$ accuracy drops to 69.45\%, compared to 72.91\% at $\gamma{=}5.0$. Small $\gamma$ values cause the SFT gradient to vanish early in training, leaving the model without sufficient target grounding. This sensitivity confirms that the SFT component plays an essential stabilizing role, consistent with our discussion in \S\ref{sec:preference} and recent findings in ORPO~\cite{orpo} and SimPO~\cite{simpo}.

\vspace{0.3cm}
\subsection{Ablation Study (\texorpdfstring{$\spadesuit$}{} RQ4).}
\label{sec:ablation}

Table~\ref{tab:ablation_study} validates every component of \method{} across all three datasets by removing one element at a time.

\vspace{0.3cm}
\paragraph{\textbf{Core components.}}
Removing self-supervised training causes the most dramatic drop, with accuracy falling from 72.91\% to 38.71\% on Citeseer and from 89.98\% to 42.15\% on Pubmed, confirming that base LLMs lack the structural inductive bias that \method{} instills. Hierarchical sampling improves accuracy by 2--3\% across datasets by focusing learning on informative 2-/3-hop distinctions rather than trivial distant pairs. Ensemble voting with $R > K$ adds another 2--3\% by averaging away anchor-sampling noise, consistent with the variance reduction effect observed in \S\ref{sec:analysis}.

\vspace{0.3cm}
\paragraph{\textbf{Loss components.}}
Among the three loss components, SFT has the largest individual impact, with accuracy dropping to 68.87\% on Citeseer and 86.33\% on Pubmed when removed. However, SFT alone is not sufficient: the adaptive $w_{\text{dist}}$ and $w_{\text{rank}}$ each contribute 1--3\% improvements, and disabling either causes a consistent drop across all three benchmarks. Notably, $w_{\text{rank}}$ has a larger effect on Cora, where the model must discriminate among 7 classes with subtle boundaries, while $w_{\text{dist}}$ matters more on Pubmed, where the 2-hop vs.\ 3-hop distinction is the primary learning signal. This confirms that the preference signal and the SFT regularizer play complementary rather than redundant roles (\S\ref{sec:preference}).
\vspace{0.1cm}
\section{Conclusion.}
\label{sec: conclusion}

This paper introduced \method{}, a self-supervised framework that trains LLMs on graph structure for few-shot node classification without any labeled training data. Our key insight is to reformulate node classification as a link prediction task, leveraging the homophily principle to turn graph topology into a free supervisory signal. Through hierarchical hop-based sampling and adaptive preference learning, \method{} learns to distinguish structurally close nodes from distant ones, implicitly capturing the class structure encoded in graph topology. Experiments on three standard TAG benchmarks demonstrate that \method{} matches fully-supervised GNNs and substantially outperforms existing graph-LLM methods across all few-shot settings, while adaptive early exit reduces inference cost by $\sim\!3.6\times$ with negligible accuracy loss.

More broadly, our results support a general claim: in homophilous graphs, topology can serve as a competitive substitute for labeled supervision when fine-tuning LLMs. This opens several promising directions for future work, including scaling to larger and denser graphs with sub-sampled or importance-weighted BFS, exploring how \method{}'s gains interact with different LLM backbones and scales, and investigating learned or structurally-informed anchor selection strategies to further improve inference reliability. We believe that structural self-supervision represents an underexplored and orthogonal direction to the current label-centric line of graph-LLM research.

\bibliographystyle{siamplain}
\bibliography{main}

\clearpage
\newpage
\appendix
\section{Scope Discussion.}
\label{app:scope}

\method{} introduces a new paradigm for graph-LLM research: using graph 
topology as a self-supervised signal to replace labeled data entirely. As 
a proof of concept for this direction, our current evaluation focuses on 
homophilous text-attributed graphs, where the homophily principle provides 
a clear and reliable preference signal. This covers a wide range of 
practical applications including citation networks, co-authorship graphs, 
and social networks. Our evaluation spans homophily ratios from 72.9\% to 
82.5\% and demonstrates robust performance within this regime. For graphs 
with strong heterophily, where edges connect nodes of different classes, 
the preference signal would need to be redefined, which we leave as a 
separate research question. Similarly, scaling to million-node graphs is 
a natural next step as the core paradigm matures.

\section{Datasets.}
\label{app:datasets}

We conduct our experiments on three widely-used citation network benchmarks: Cora, Citeseer, and Pubmed. These datasets are standard in the evaluation of graph-based node classification methods. Each node represents a scientific publication, and edges represent citation links between them. The task is to classify each publication into its respective academic field based on its textual content (title and abstract) and its position within the citation network.

Following the setup of recent comprehensive benchmarks~\cite{wu2025llms, xu2025gnn}, we utilize versions of these datasets where the raw text for each node is readily available. This ensures that all methods, including our LLM-based framework and the baselines, have access to the same rich semantic information. The key statistics for these datasets are summarized in Table~\ref{tab:dataset_stats_concise}.

\begin{table}[h!]
\centering
\caption{Summary statistics of the evaluation datasets. Homophily Ratio is the fraction of edges connecting nodes of the same class.}
\label{tab:dataset_stats_concise}
\resizebox{0.48\textwidth}{!}{
\begin{tabular}{lcccc}
\toprule
\textbf{Dataset} & \textbf{Nodes} & \textbf{Edges} & \textbf{Classes} & \textbf{Homophily Ratio (\%)} \\
\midrule
Cora      & 2,708 & 5,278  & 7  & 82.5 \\
Citeseer  & 3,327 & 4,676  & 6  & 72.9 \\
Pubmed    & 19,717& 44,324 & 3  & 79.2 \\
\bottomrule
\end{tabular}}
\end{table}
Note that all three datasets exhibit high homophily ratios (>70\%), validating the core assumption of our approach that graph topology encodes class structure through connection patterns.

A brief description of each dataset is provided below:
\begin{itemize}[leftmargin=*, itemsep=0pt, topsep=0pt]
    \item \textbf{Cora}~\cite{sen2008collective}: The Cora dataset consists 
    of machine learning papers categorized into seven classes, including 
    \textit{Neural Networks}, \textit{Reinforcement Learning}, and 
    \textit{Probabilistic Methods}. Cora is a relatively small graph but 
    exhibits high homophily (82.5\%), making it a standard benchmark for 
    methods that leverage neighborhood similarity.
    \item \textbf{Citeseer}~\cite{giles1998citeseer}: The Citeseer dataset 
    contains scientific publications classified into six categories covering 
    areas such as \textit{Artificial Intelligence}, \textit{Machine Learning}, 
    and \textit{Human-Computer Interaction}. Compared to Cora, Citeseer has 
    a slightly lower homophily ratio (72.9\%), presenting a more challenging 
    scenario where cross-class connections are more frequent.
    \item \textbf{Pubmed}~\cite{yang2016revisiting}: The Pubmed dataset 
    comprises biomedical articles from the PubMed database concerning diabetes, 
    classified into three categories: \textit{Experimentally induced diabetes}, 
    \textit{Type 1 diabetes}, and \textit{Type 2 diabetes}. It is a larger 
    and denser graph with high homophily (79.2\%). Its distinct biomedical 
    domain makes it valuable for testing generalization beyond computer 
    science literature.
\end{itemize}

\textbf{Few-shot evaluation protocol.} For all experiments, we follow the standard few-shot evaluation setup where $K \in \{5, 10, 20\}$ labeled nodes per class are available. Crucially, for our \method{}, these labels are only used during inference as anchor nodes for comparison, not during training. For baseline methods that require supervised training (e.g., \llmname{InstructTuning}, \llmname{LLaGA}, \llmname{GraphGPT}), the same $K$ labeled nodes per class are used for fine-tuning. We follow the data splits provided by~\cite{wu2025llms, xu2025gnn} to ensure fair comparison.

\section{Baseline Methods.}
\label{app:baselines}

We compare \method{} against two categories of baselines: (1) traditional Graph Neural Networks (GNNs) that operate on node features and graph structure, and (2) recent LLM-based approaches that leverage text understanding for node classification. Below we describe each baseline and its implementation details.

\subsection{Non-LLM Baselines.}

This category includes representative GNN architectures that are widely 
considered strong baselines for node classification:

\begin{itemize}[leftmargin=*, itemsep=0pt, topsep=0pt]
    \item \textbf{\llmname{GCN} (Graph Convolutional Network)}~\cite{kipf2016semi}: 
    A foundational GNN model that learns node representations by aggregating 
    information from immediate neighbors through spectral graph convolution.
    \item \textbf{\llmname{GraphSAGE} (Graph Sample and Aggregate)}~\cite{hamilton2017inductive}: 
    An inductive framework that generates node embeddings by sampling neighbors 
    and applying aggregation functions (e.g., mean, max-pooling).
    \item \textbf{\llmname{SGC} (Simple Graph Convolution)}~\cite{wu2019simplifying}: 
    A simplified variant of \llmname{GCN} that removes non-linear activations between 
    layers, reducing complexity while maintaining competitive performance 
    on homophilous graphs.
    \item \textbf{\llmname{GAT} (Graph Attention Network)}~\cite{velickovic2017graph}: 
    A GNN that incorporates attention mechanisms, allowing nodes to assign 
    different weights to neighbors during aggregation.
\end{itemize}

\paragraph{Implementation Details.}
Our GNN baselines are implemented using PyTorch Geometric~\cite{fey2019fast}. For the few-shot evaluation, each GNN is trained in a semi-supervised (transductive) setting. This means the model is trained using only the labels of the $K$ nodes per class from the training set, but it has access to the full graph structure and the features of all nodes during training. We adopt a consistent set of hyperparameters for all GNN models to ensure comparability. We use the Adam optimizer~\cite{kingma2014adam} with a learning rate of \texttt{1e-2} and a weight decay of \texttt{5e-4}. Models are trained for a maximum of 500 epochs with an early stopping mechanism that has a patience of 100 epochs, monitored on a validation set. Following common practice, the hidden dimension for all GNN models is set to 128, and we generally use a 2-layer architecture with a dropout rate of 0.5.

\subsection{LLM-based Baselines.}
\label{app:llm_baselines}

This category includes methods that leverage LLMs for node classification, 
all using \texttt{LLaMA-3-8B-Instruct} as the backbone for fair comparison. 
These baselines cover different paradigms of integrating LLMs with graph data:

\begin{itemize}[leftmargin=*, itemsep=0pt, topsep=0pt]
    \item \textbf{Zero-Shot}: The LLM classifies a node based solely on 
    its textual attributes (title and abstract) without any examples. The 
    prompt provides the node's text and asks the model to choose from the 
    list of possible class labels.
    
    \item \textbf{w/ Neighbors}: An enhanced zero-shot baseline that 
    incorporates structural information by including the titles of 1-hop 
    neighbors in the prompt, providing local graph context.
    
    \item \textbf{w/ Few Shot (In-Context Learning)}: This baseline performs 
    in-context learning by providing demonstration examples in the prompt. 
    However, due to context length constraints, we cannot include all $K$ 
    examples per class (which would be $K \times C$ total examples) in a 
    single prompt. Instead, we employ a voting strategy similar to our inference 
    approach: in each round, we randomly sample one demonstration example per 
    class (without replacement from the pool of $K$ examples), present them 
    to the LLM along with the query node, and collect the prediction. We 
    repeat this process $K$ times with different example combinations, and 
    determine the final classification via majority voting. This approach 
    enables the model to learn from multiple diverse example combinations 
    while respecting context length limitations. The demonstrations are formatted 
    as: \texttt{"Example: [text] $\rightarrow$ Answer: [label]"} before the 
    query.

    \item \textbf{InstructTuning}~\cite{ouyang2022training}: Standard 
    supervised fine-tuning of \texttt{LLaMA-3-8B-Instruct} on the $K$ 
    labeled training nodes per class. We employ Parameter-Efficient Fine-Tuning 
    (PEFT) using Low-Rank Adaptation (LoRA)~\cite{hu2022lora} with rank 
    8 and alpha 16. Training data is formatted as instruction-response pairs.

    \item \textbf{LLaGA}~\cite{chen2024llaga}: A state-of-the-art method 
    that uses a GNN to create structure-aware node representations, which 
    are then projected and fed into the LLM along with node text. Only the 
    projection layer is trained, making it parameter-efficient. We follow 
    their official implementation and training protocol.
    
    \item \textbf{GraphGPT}~\cite{tang2024graphgpt}: A graph-specific 
    instruction-tuning method that employs multi-stage training to align 
    textual semantics with graph structure. We follow the official implementation 
    for fine-tuning on the downstream classification task.
\end{itemize}

\paragraph{Prompt Design.} For zero-shot and few-shot baselines, we employ 
a consistent instruction prompt template shown below. The template uses 
placeholders that are replaced with dataset-specific information: 
\texttt{\{raw\_text\}} contains the node's title and abstract, 
\texttt{\{num\_classes\}} specifies the number of classes, and 
\texttt{\{labels\}} lists the class names.

\begin{tcolorbox}[
  colback=myblue!5!white,      
  colframe=myblue!85!black,    
  fonttitle=\bfseries\sffamily, 
  title=General Node Classification Prompt
]
\small 

\textbf{\texttt{Instruction:}} \texttt{Given the following text from a scientific paper, classify it into one of the} \textcolor{myred}{\texttt{\{num\_classes\}}} \texttt{possible categories. The paper's text is:}

\textcolor{mybrown}{\texttt{\{raw\_text\}}}

\textbf{\texttt{The possible categories are:}} \textcolor{mossgreen}{\texttt{\{labels\}}.}

\textbf{\texttt{Question:}} \texttt{Which category does this paper belong to? Please respond with only the category name.}

\textbf{\texttt{Answer:}}
\end{tcolorbox}

For the \textbf{w/ Few Shot} baseline, we prepend $K$ examples per class 
before the query in the format: \texttt{"Example [i]: [text] $\rightarrow$ 
Answer: [label]"}. For the \textbf{w/ Neighbors} baseline, we append neighbor 
information after the main text: \texttt{"This paper cites/is cited by: 
[neighbor titles]"}.

\section{\method{} Training Configuration.}
\label{app:hoprank_config}

We provide detailed hyperparameters and training configurations for \method{} 
to ensure reproducibility.

\paragraph{Model Architecture.} For fair comparison, we use \texttt{LLaMA-3-8B-Instruct} as 
the base model with LoRA~\cite{hu2022lora} for parameter-efficient fine-tuning. 
LoRA is applied to all linear layers (\texttt{lora\_target="all"}) with 
rank 8 and alpha 16. This results in approximately 0.5\% trainable parameters 
relative to the full model, significantly reducing memory requirements while 
maintaining performance.

\paragraph{\method Loss Configuration.} Our loss function combines three 
components as described in Eq. (\ref{eq:hoprank}):
\begin{itemize}[leftmargin=*, noitemsep, topsep=0pt]
    \item \textbf{DPO temperature} ($\beta$): 0.1, controlling the strength 
    of preference optimization
    \item \textbf{SFT weight} ($\gamma$): 5.0, balancing 
    preference learning with supervised fine-tuning
\end{itemize}

\paragraph{Training Hyperparameters.} The complete training configuration is:
\begin{itemize}[leftmargin=*, noitemsep, topsep=0pt]
    \item \textbf{Optimizer}: AdamW with learning rate 2e-4
    \item \textbf{Learning rate schedule}: Cosine annealing with 10\% warmup
    \item \textbf{Batch size}: 2 per device with 4 gradient accumulation 
    steps (effective batch size = 8)
    \item \textbf{Training epochs}: 10 epochs (sufficient for convergence 
    with preference learning)
    \item \textbf{Max sequence length}: 4096 tokens
    \item \textbf{Gradient clipping}: 1.0 (default)
\end{itemize}

\paragraph{Hierarchical Sampling Details.} For each training edge, we 
sample one negative from each of \{2-hop, 3-hop\} neighborhoods, 
creating instances with 1 chosen response and 2 rejected responses. Nodes 
at each hop distance are sampled uniformly at random. If a particular hop 
neighborhood is empty, we skip it and only use available hop distances. 

\paragraph{Evaluation and Model Selection.} We evaluate every 20 training 
steps and employ early stopping with patience of 8 evaluations (160 training 
steps) based on evaluation loss. The best model checkpoint is saved and 
used for inference. Training typically converges within 2-3 epochs across 
all datasets.

\paragraph{Inference Configuration.} During few-shot inference:
\begin{itemize}[leftmargin=*, noitemsep, topsep=0pt]
    \item \textbf{Anchor selection}: Random sampling with replacement when 
    $R > K$
    \item \textbf{Voting rounds}: $R = 100$ for all experiments
    \item \textbf{Decoding}: Sampling with temperature 0.8
\end{itemize}

The ensemble sampling strategy with $R = 100$ provides a good balance between 
accuracy and computational cost, allowing each anchor to be evaluated 
approximately more times on average across different contexts.

\paragraph{Computational Resources.} All experiments are conducted on 2 NVIDIA 
H100 GPUs (80GB). Training time varies by dataset: Cora and Citeseer complete 
in approximately 1-2 hours, while Pubmed (larger graph) takes 3-4 hours.

\section{Training Data Format and Case Study.}
\label{app:case_study}

To illustrate the hierarchical hop-based sampling approach, we provide a 
complete example showing the exact prompt format used during training. The 
prompt structure follows the same format as shown in Section~\ref{sec:sampling}, with 
abstracts truncated here for brevity.

\paragraph{Complete Training Prompt Example.} Below is a representative 
training instance from the Cora dataset:
\begin{table*}[t!]
\centering
\caption{Training prompt example from the Cora dataset.}
\label{tab:train_prompt_example}
\begin{tcolorbox}[
  colback=myblue!5!white,
  colframe=myblue!85!black,
  fonttitle=\bfseries,
  title=Training Prompt Example from Cora Dataset,
  width=\textwidth,
]
\small

\textbf{Given a source paper and multiple candidate papers, identify which 
candidate paper is most likely to have a citation relationship with the 
source paper.}

\textbf{Source Paper:} `Statistical Ideas for Selecting Network Architectures'

\textbf{Abstract:} Choosing the architecture of a neural network is one 
of the most important problems in making neural networks practically useful...

\textbf{Candidate Papers:}

\textbf{A.} `Reasoning about Time and Probability'

\textbf{Abstract:} An essential component of an intelligent agent is the 
ability to notice, encode, store, and utilize information about...

\textcolor{mossgreen}{[3-hop: easy negative]}

\textbf{B.} `Bayesian Graphical Models for Discrete Data'

\textbf{Abstract:} York's research was supported by a NSF graduate fellowship. 
The authors are grateful to Julian Besag, David Bradshaw...

\textcolor{myred}{[1-hop: actual neighbor]}

\textbf{C.} `Model Selection for Generalized Linear Models via GLIB, with 
Application to Epidemiology'

\textbf{Abstract:} This is the first draft of a chapter for Bayesian 
Biostatistics, edited by Donald A. Berry and Darlene K. Strangl...

\textcolor{mybrown}{[2-hop: hard negative]}

\textbf{Which candidate paper is most likely to be cited by or cite the 
source paper? Provide your answer by reproducing the title and abstract 
of the selected paper.}

\hrulefill

\textbf{Chosen Response (Ground Truth):}

Bayesian Graphical Models for Discrete Data

\textbf{Rejected Responses:}
\begin{itemize}[leftmargin=1em, itemsep=0pt, parsep=0pt]
    \item Model Selection for Generalized Linear Models via GLIB, with 
    Application to Epidemiology \quad \texttt{[2-hop]}
    \item Reasoning about Time and Probability \quad \texttt{[3-hop]}
\end{itemize}

\end{tcolorbox}
\end{table*}

\paragraph{Preference Structure.} This instance creates three pairwise 
comparisons for the \method{}Loss:
\begin{itemize}[leftmargin=*, itemsep=0pt, topsep=0pt]
    \item \textbf{Chosen vs. 2-hop negative} (Paper B vs. Paper C): Both 
    papers discuss statistical model selection, making this a challenging 
    distinction. Distance weight: $w_{\text{dist}} = 1/(2-1) = 1.0$ (highest).
    
    \item \textbf{Chosen vs. 3-hop negative} (Paper B vs. Paper A): Semantic 
    topics differ significantly (Bayesian graphical models vs. agent planning). 
    Distance weight: $w_{\text{dist}} = 1/(3-1) = 0.5$ (lower).
\end{itemize}

The ranking weight $w_{\text{rank}}$ adapts during training based on the 
model's current ranking of these candidates, emphasizing pairs where ranking 
errors occur at top positions.


\end{document}